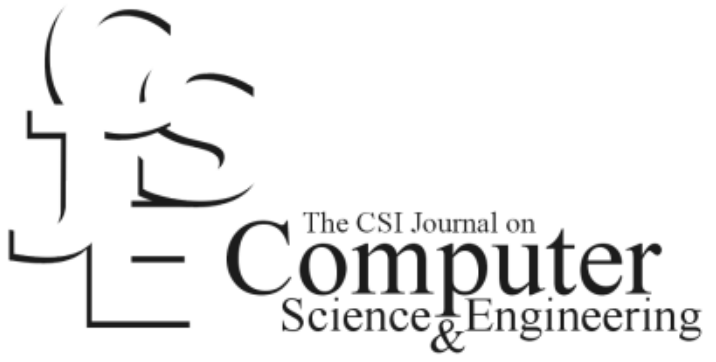

# Comparing Time-Series Analysis Approaches Utilized in Research Papers to Forecast COVID-19 Cases in Africa: A Literature Review

Ali Ebadi[1] Ebrahim Sahafizadeh[1*]

[1] Department of Computer Engineering, Persian Gulf University, Bushehr, Iran
* Corresponding Author Email: sahafizadeh@pgu.ac.ir

## Abstract

This literature review aimed to compare various time-series analysis approaches utilized in forecasting COVID-19 cases in Africa. The study involved a methodical search for English-language research papers published between January 2020 and July 2023, focusing specifically on papers that utilized time-series analysis approaches on COVID-19 datasets in Africa. A variety of databases, including PubMed, Google Scholar, Scopus, and Web of Science, were utilized for this process. The research papers underwent an evaluation process to extract relevant information regarding the implementation and performance of the time-series analysis models. The study highlighted the different methodologies employed, evaluating their effectiveness and limitations in forecasting the spread of the virus. The result of this review could contribute deeper insights into the field, and future research should consider these insights to improve time series analysis models and explore the integration of different approaches for enhanced public health decision-making.

## 1. Introduction

The devastating impact of the COVID-19 pandemic is globally significant, striking Africa with around 10 million reported cases and over 175,000 deaths by July 2023 [1]. This unprecedented health crisis has spurred researchers worldwide to investigate COVID-19 datasets, with a specific focus on Africa, utilizing various time-series analysis methods with the driving aim of modelling and forecasting the spread of the virus, and identifying contributing factors to its transmission.

In this systematic literature review, our objective is to examine and compare these time-series analysis methods featured in COVID-19 research papers, with datasets specific to Africa. The review not only sets out to present an overview of the methodologies' current application but also intends to identify existing gaps in knowledge that may facilitate refined research in the future.

Ultimately, this critical analysis strives to shed light on the comparative efficacy of different time-series analysis methods, such as ARIMA, LSTM, and others, contributing to a comprehensive understanding of their application in studying and predicting the spread of COVID-19 in Africa. This understanding, we believe, will prove invaluable in improving our preparedness and response to future pandemics.

The extensive search for this review comprised academic databases like PubMed, Google Scholar, Scopus, and Web of Science, scanning research papers published between January 2020 and July 2023 that utilized time-series analysis on COVID-19 datasets in Africa. The review sought to engage the areas for improvement and standardization in the field, underlining the value of a comprehensive and accessible database for COVID-19 data specific to Africa, and indicating the role of sophisticated predictive tools for future mitigation efforts.

This could lead to a more robust strategy that balances the necessity of effective forecasting with the demand for practical, data-informed strategies to combat the viral spread in Africa, specifically, and worldwide in general. Our findings and recommendations will serve as critical input for researchers and policymakers in their quest to refine existing models or design innovative methodologies for disease surveillance and management.

The rest of the paper is organized as follows: Section 2 contains methods and materials. Section 3 focuses on the forecasting models utilized in studies. Section 4 presents the results obtained from research papers, and Section 5 is the discussion section.

A preprint version of this work has been made available on *arXiv* (arXiv: 2310.03606).

# 2. Methods and Materials

## 2.1. Information Sources and Search

We conducted a comprehensive search of academic databases, including PubMed, Google Scholar, Scopus, and Web of Science, for studies published between January 2020 and July 2023. The search strategy targeted research articles that applied time-series analysis approaches to COVID-19 datasets in African countries. Keywords and combinations included: "COVID-19," "coronavirus," "time-series analysis," "forecasting models," "statistical methods," "Africa," "ARIMA," "LSTM," "RNN," "machine learning," and "deep learning."

## 2.2. Selection Criteria

We included research papers that satisfied the following criteria: (a) applied time-series analysis or

Table 1.

## 2.3 Study Selection Process

The study selection process followed the PRISMA 2020 guidelines. A total of 160 records were identified from databases. After removal of 30 duplicates, 130 records were screened by title and abstract, with 100 excluded. Thirty full-text reports were retrieved, of which 2 were unavailable. Twenty-eight articles were assessed for eligibility, and 18 were excluded for the following reasons: not focused on Africa (n = 3), not peer-reviewed or not in English (n = 8), and lacking both forecasts of daily or cumulative confirmed cases and model performance metrics (n = 7). Ultimately, 10 studies met all inclusion criteria and were included in this review. The full process is illustrated in the PRISMA 2020 flow diagram (Figure 1)[2].

## 2.4. Data Extraction

We extracted the following information from each research paper: (1) the title and authors; (2) the research question or aim of the study; (3) the time-series analysis forecasting approaches on COVID-19 datasets in Africa; (b) published in English language peer-reviewed journals or conference proceedings; and (c) published between January 2020 and July 2023. Studies were excluded if they did not report forecasts of daily or cumulative confirmed cases and corresponding model performance metrics, or if they did not meet the population, study type, or publication requirements. The full set of inclusion and exclusion criteria is summarized in

# 3. Overview of Forecasting Models

Various time-series analysis approaches have been utilized to predict the spread of COVID-19.

## 3.1. Autoregressive Integrated Moving Average (ARIMA)

ARIMA (Autoregressive Integrated Moving Average) is a popular time series forecasting method used in various fields such as finance, economics, and engineering. It models the underlying patterns and trends in a time series data to predict future values based on its past values.

ARIMA model is composed of three essential elements: autoregression (AR), differencing (I), and moving average (MA). The autoregression component (AR) captures the connection between the current value of the time series and its previous values. It forecasts the next value of the time series by leveraging the preceding values. The moving average component (MA) focuses on the relationship between the current value and the past errors. It predicts the forthcoming value of the time series based on prior errors. Finally, the (I) component deals with the non-stationarity of the time series by taking differences between consecutive observations.

approach utilized; (4) the data sources used; (5) the reported daily confirmed cases or cumulative number of confirmed cases; (6) the performance metrics used to evaluate the model; (7) limitations of models; and (8) any other relevant information.

# 3. Overview of Forecasting Models

Various time-series analysis approaches have been utilized to predict the spread of COVID-19.

## 3.1. Autoregressive Integrated Moving Average (ARIMA)

ARIMA (Autoregressive Integrated Moving Average) is a popular time series forecasting method used in various fields such as finance, economics, and engineering. It models the underlying patterns and trends in a time series data to predict future values based on its past values.

ARIMA model is composed of three essential elements: autoregression (AR), differencing (I), and moving average (MA). The autoregression component (AR) captures the connection between the current value of the time series and its previous values. It forecasts the next value of the time series by leveraging the preceding values. The moving average component (MA) focuses on the relationship between the current value and the past errors. It predicts the forthcoming value of the time series based on prior errors. Finally, the (I) component deals with the non-stationarity of the time series by taking differences between consecutive observations.

Table 1. Inclusion and Exclusion Criteria for Study Selection

| Domain | Inclusion Criteria | Exclusion Criteria |
|---|---|---|
| Population/Setting | Studies focusing on COVID-19 forecasting in African countries or regions | Studies not involving African countries; global studies without Africa-specific results |
| Model Type | Time-series or machine/deep learning approaches | Studies with no forecasting component (e.g., descriptive epidemiology only) |
| Outcome & Metrics | Reported forecasts of daily or cumulative confirmed cases and provided at least one quantitative accuracy metric | Studies that did not report forecasts of daily or cumulative confirmed cases and did not provide performance metrics |
| Publication Type | Peer-reviewed journal articles, published in English | Non-peer-reviewed |
| Time Window | Published between January 2020 and July 2023 | Published outside the search period |

The ARIMA model is denoted as ARIMA (p, d, q), where p indicates the AR component's order, d represents the differencing order, and q signifies the MA component's order. The formula for an ARIMA (p, d, q) model can be written as:

$$x_t = c + a_1 x_{(t-1)} + \ldots + a_p x_{(t-p)} + z_t + y_1 z_{(t-1)} + \ldots + y_q z_{(t-q)} \quad (1)$$

where $x_t$ is the value of the time series at time $t$, $c$ is a constant, $a_i$ and $y_i$ are the parameters of the AR and MA components, respectively, and $z_t$ is the error term at time $t$.

In addition to ARIMA model, SARIMA (Seasonal Autoregressive Integrated Moving Average) model was employed for predicting COVID-19 patterns. SARIMA model includes both ARIMA parameters (p, d, q) and seasonal terms $(P, D, Q)_m$, where P represents the seasonal autoregressive term, D denotes the seasonal differencing term, Q signifies the seasonal moving average term, and m indicates the number of observations per year. The SARIMA model is mathematically represented as follows:

$$\acute{E}_{,P} (B^m)\, \phi_p (B)\, (1 - B^m)^D\, (1 - B)^d\, y_t = \Theta_Q (B^m)\, \theta_q (B)\, w_t \quad (2)$$

where $y_t$ is the non-stationary time-series, $w_t$ is the Gaussian white noise process, $\acute{E}_{,P} (B^m)$ is seasonal moving average polynomial, $\Theta_Q (B^m)$ is a seasonal moving average polynomial, and $B$ is a backshift operator [3].

The ARIMA model is a commonly used tool for predicting COVID-19 trends, but it has limitations that can affect its effectiveness. One limitation is its assumption of stationarity, which may not hold true for COVID-19 data. Another limitation is its linearity, which may not capture complex non-linear relationships in the data. The accuracy of ARIMA forecasting also depends on the quality of input data and assumes that future behavior is solely dependent on past behavior. However, COVID-19 is a complex phenomenon affected by many factors.

## 3.2. Long short-term memory (LSTM)

LSTM-based model is a powerful extension of artificial recurrent neural networks (RNNs) that effectively addresses the vanishing gradient problem. This model enhances the RNNs' memory capacity, allowing them to learn and retain long-term dependencies of inputs. The LSTM memory cell is known as a "*gated*" cell, where the term "*gate*" refers to the ability to decide whether to keep or discard the memory information. This extension enables the LSTM model to read, write, and delete information from their memories over an extended period.

By capturing important features from inputs and preserving this information over time, an LSTM model can decide which information is worth preserving or discarding, based on the weight values assigned during the training process. As a result, an LSTM model can learn to differentiate between valuable and irrelevant information and make informed decisions based on this input.

Typically, an LSTM model is composed of three gates: the forget gate, input gate, and output gate. The forget gate is responsible for deciding whether to keep or discard the current information, while the input gate determines the amount of new information to be added to the memory. Lastly, the output gate controls how much of the current value in the cell should be included in the output.

*1) Forget Gate*. It uses a sigmoid function to determine which information should be discarded from the LSTM memory. This decision is based on the values of the previous hidden state ($h_{t-1}$) and the current input ($x_t$). The output of the Forget Gate is denoted as $f_t$, and is a value between 0 and 1. A value of 0 indicates that the learned information should be completely discarded, while a value of 1 indicates that the information should be preserved in its entirety. The output of the Forget Gate is computed using the following equation:

$$f_t = \sigma (W_{fh} [h_{t-1}], W_{fx} [x_t], b_f) \quad (3)$$

where $b_f$ is a constant and is called the bias value.

*2) Input Gate.* Its role is to determine whether new information should be incorporated into the LSTM memory or not. This gate is composed of two layers: a sigmoid layer and a "tanh" layer. The sigmoid layer identifies which values should be updated, while the "tanh" layer produces a vector of potential new values that can be added to the LSTM memory. The outputs of these two layers are computed as follows:

$$i_t = \sigma (W_{ih} [h_{t-1}], W_{ix} [x_t], b_i) \quad (4)$$

$$\hat{c}_t = \tanh (W_{ch} [h_{t-1}], W_{cx} [x_t], b_c) \quad (5)$$

in which it indicates whether the value needs to be updated or not, and ĉt represents a vector of new candidate values that will be added into the LSTM memory. By combining these two layers, the LSTM memory undergoes an update process. The forget gate layer plays a role in discarding the current value (ct−1) by multiplying it with the old value, while the new candidate value (it * ĉt) is added to the memory. The following equation represents its mathematical equation:

$$c_t = f_t * c_{t-1} + i_t * \hat{c}_t \quad (6)$$

where $f_t$ is the result of the forget gate, which is a value between 0 and 1, where 0 indicates completely get rid of the value; whereas, 1 implies completely preserve the value.

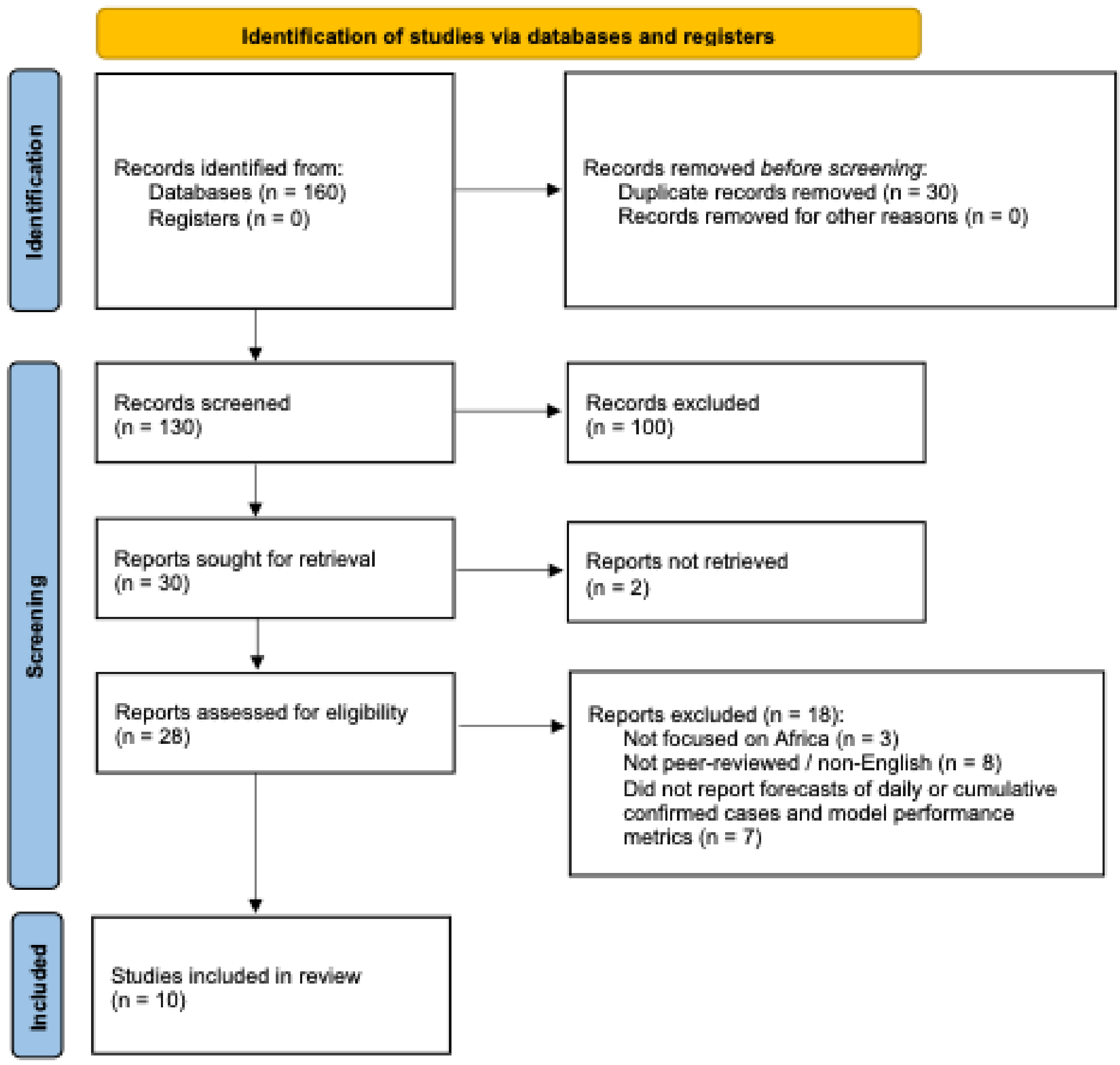

**Figure 1. PRISMA 2020 flow diagram of study selection.** From 160 database records, 28 full-texts were assessed, and 10 studies met the inclusion criteria for this review.

*3) Output Gate.* To generate the output, this gate initially employs a sigmoid layer to identify the relevant portions of the LSTM memory. Subsequently, it applies a non-linear "tanh" function to transform the obtained values into a range between -1 and 1. Finally, the output of the "tanh" layer is multiplied by the output of another sigmoid layer. This gating mechanism enables the LSTM model to selectively output relevant information while suppressing irrelevant information. The computation of the output of the LSTM model can be represented by the following equation:

$$o_t = \sigma\left(W_{oh}[h_{t-1}], W_{ox}[x_t], b_o\right). \quad (7)$$

$$h_t = o_t * \tanh(c_t) \quad (8)$$

where $o_t$ is the output value, and $h_t$ is its representation as a value between −1 and 1 [4].

Overall, these three gates work together to manage the flow of information within the LSTM model. By selectively retaining and discarding information, an LSTM model can effectively learn long-term dependencies and make accurate predictions based on the input data.

The LSTM model has demonstrated promising results in forecasting COVID-19 trends in Africa. However, the model also has limitations that can affect its effectiveness in predicting COVID-19 trends. One challenge is that the model is computationally complex, which can make it difficult to train, particularly with large datasets. Moreover, the LSTM model requires a substantial amount of training data to reach optimal performance and is susceptible to overfitting. Additionally, the model only considers past COVID-19 case trends and may not account for external factors such as government policies, public health interventions, and social behaviors that can influence future trends.

## 3.3. Gated Recurrent Unit (GRU)

The Gated Recurrent Unit (GRU) is an advanced recurrent neural network (RNN) architecture, introduced by Cho et al. [5], designed to overcome the limitations of traditional RNNs. GRU utilizes gating mechanisms to effectively handle sequential data. Although GRU and LSTM share similar motivations, they differ in architecture and computational complexity. GRU has a simpler architecture with two gates (update and reset), sacrificing some memory capacity but offering computational efficiency. The update gate manages the flow of information from the previous hidden state to the current time step. It determines how much of the past information should be retained and how much new information should be incorporated. The reset gate controls the amount of past information that is discarded. By combining the previous hidden state with the output of the reset gate, a current memory is created, capturing relevant information from the past. The hidden state is computed by combining this current memory with the output of the update gate. It serves as both the output of the GRU at a specific time step and the input for the next time step, enabling the model to capture relationships across the sequence.

The computations in a GRU can be expressed through the following equations [6]:

1) *Hidden State:*

$$h_t = (1 - z_t) * h_{t-1} + z_t * \hat{h}_t \tag{9}$$

2) *Update Gate:*

$$z_t = \sigma (W_z * [h_{t-1}, x_t]) \tag{10}$$

3) *Reset Gate:*

$$r_t = \sigma (W_r * [h_{t-1}, x_t]). \tag{11}$$

4) *Candidate Activation:*

$$\hat{h}_t = \tanh (W * [r_t * h_{t-1}, x_t]) \tag{12}$$

GRU model has limitations that can impact their effectiveness. One limitation is their sensitivity to initialization values, which can cause the model to converge to suboptimal solutions. Additionally, GRU model requires a large amount of training data to achieve optimal performance and can be prone to overfitting. Another challenge is the model's complexity, which can make it computationally expensive to train and deploy, particularly with large datasets. Moreover, just like LSTM model, GRU model only considers past trends in COVID-19 cases and may not incorporate external factors that influence future trends.

## 3.4. Alpha-Sutte Method

The Alpha-Sutte method was created for predicting finance, insurance, and time-series data. It has its roots in the Sutte Indicator [7]. The Alpha-Sutte Indicator approach is a newly invented strategy. When making predictions, the Alpha-Sutte Indicator utilizes the preceding four data points and does not necessitate any assumption tests. As a result, it offers greater flexibility for application to diverse datasets [8].. It is especially well-suited for making short-term predictions of data series. The formula of the Alpha-Sutte method is shown below [9]:

$$a_t = \frac{\alpha(\Delta x/((\alpha+\delta)/2)) + \beta(\Delta y/((\alpha+\beta)/2)) + \gamma(\Delta z/((\beta+\gamma)/2))}{3} \tag{13}$$

where $\delta = a_{t-4}$, $\alpha = a_{t-3}$, $\beta = a_{t-2}$, and $\gamma = a_{t-1}$.

$$\Delta x = \alpha - \delta,\ \Delta y = \beta - \alpha,\ \Delta z = \gamma - \beta, \tag{14}$$

$a_t$ is the observation at time $t$.

While Alpha-Sutte Indicator model only requires a small amount of data to generate accurate forecasts, this can make it less suitable for longer-term forecasting or modelling more complex phenomena.

To assess the accuracy of the forecasted results in the papers, we employed Root Mean Square Error (RMSE). RMSE is a commonly used metric to measure the accuracy of a forecast. It provides an overall assessment of the model's prediction error by calculating the square root of the average of the squared differences between the predicted values and the actual values. The formula for calculating RMSE is as follows:

$$RMSE = \sqrt{\sum_{i=1}^{n} \frac{(y_i - y)^2}{n}} \tag{15}$$

where $y$ is the predicted value from the model and $y_i$ is the actual value of the data point.

# 4. Results

In this section, we gathered valuable information from each study together. We considered only daily confirmed cases or cumulative confirmed cases. For studies that did not report RMSE values, we calculated RMSE using the corresponding time series from Our World in Data [10]; these recalculated values are clearly marked in Table 2. We note that some studies originally relied on different datasets, and therefore, the reported RMSEs in this review may not be directly comparable to the values in the original publications. An overview of forecasts from the included studies is presented in Figure 2. Subfigures (a–s) illustrate reconstructed comparisons between actual and predicted cases wherever sufficient information was available. For studies where forecast plots were already reported in the original publications, only the results are summarized in text and not re-visualized here.

In two separate studies conducted by K.E. ArunKumar et al. [3], [11], the authors focused on forecasting COVID-19 trends for ten countries, including South Africa. The studies employed both deep learning and statistical models to forecast the cumulative confirmed, recovered, and death cases of COVID-19. However, for the purpose of this study, only the cumulative confirmed cases were considered.

The first study [11] presents utilizing GRU and LSTM models for a 60-day forecast until the end of September 2020. The RMSE values for GRU and LSTM were reported as 37,428 and 2,428, respectively. The reconstructed forecasts for South Africa from this study are shown in Figure 2a.

The comparison between LSTM and GRU was the goal of another study too. Nahla F. Omran et al. [12] presented a comparative study on the application of deep learning methods for forecasting COVID-19 cases in Egypt, Saudi Arabia, and Kuwait. They employed LSTM and GRU to forecast the confirmed cases and death cases of COVID-19. The dataset used in the study is the Novel Corona Virus 2019 Dataset obtained from Kaggle. It includes daily-level information on the number of confirmed cases and deaths of coronavirus in different countries. The time series data covers the period from January 5, 2020, to June 12, 2020. The authors presented the confirmed cases of COVID-19 for 220 days and the results of LSTM and GRU for the confirmed cases in Egypt. Among the GRU variations, the single-layer GRU demonstrated the best performance with an RMSE of 670.30478. On the other hand, the two-layer GRU exhibited the poorest performance with an RMSE of 5081.1000. Additionally, the second-best performance was observed with the single-layer LSTM, which yielded an RMSE of 1067.02289.

COVID-19 in Egypt was analyzed in other studies too. Mohamed Marzouk et al. [13] focused on using artificial intelligence-based models to predict the prevalence of the COVID-19 outbreak in Egypt. The models employed in the study include LSTM, convolutional neural network, and multilayer perceptron neural network. The models were trained and validated using data from February 14, 2020, to August 15, 2020. The LSTM model demonstrated the best performance in forecasting cumulative infections for one week and one month ahead. The researchers then applied the LSTM model with optimal parameter values to forecast the spread of the epidemic for one month ahead, using data from February 14, 2020, to June 30, 2021. We evaluated the accuracy of the model, and the RMSE was around 1,724. The forecast curves for Egypt are illustrated in Figure 2b.

Adewale F. Lukman et al. [14] focused on monitoring and predicting the prevalence of COVID-19 in South Africa, Egypt, Nigeria, and Ghana. The study utilizes the ARIMA models to forecast the trend of COVID-19 in these countries. The dataset used in the analysis covers the period from February 21, 2020, to June 16, 2020, and was extracted from the World Health Organization website. The chosen ARIMA models with statistically significant parameters and minimum Akaike information criterion correction (AICc) were ARIMA (0,2,3), ARIMA (0,1,1), ARIMA (3,1,0), and ARIMA (0,1,2) for South Africa, Nigeria, Ghana, and Egypt, respectively. They were used to make a daily forecast from 16th June to July 5, 2020. We evaluated the accuracy of the models. The RMSE values for South Africa, Nigeria, Ghana, and Egypt were around 727.190, 50.082, 122.331, and 293.516, respectively. Figure 2c–f show the forecasts for each of these countries.

In the next study by K.E. ArunKumar et al. [3], the objective was to forecast COVID-19 trends using both deep learning and statistical models. The forecast period covered 60 days until August 22nd, 2021. The deep learning models employed were GRU and LSTM, while the statistical techniques included ARIMA and SARIMA. To evaluate the performance of the models, the RMSE values were calculated by comparing the forecasted values generated by each model with the actual values for South Africa on the forecasted date. The results indicated that the ARIMA model exhibited the best performance among the models assessed. Following ARIMA, the performance rankings were as follows: LSTM, GRU, and SARIMA. The reported RMSE values for GRU, LSTM, ARIMA, and SARIMA were approximately 230,225, 180,225, 80,225, and 319,775, respectively. The comparison of these models for South Africa is presented in Figure 2g.

Despite the better performance of ARIMA compared to LSTM in [3], in a study by Mohamed Amine Rguibi et al. [15], LSTM performed better. In this study, real data were collected on COVID-19 transmission in Morocco and were used to train the ARIMA and LSTM models. The researchers then used these models to forecast the number of confirmed and death cases from 22 November to 21 January 2021. The RMSE values for the number of confirmed cases for ARIMA and LSTM were 1862.109 and 795.293, respectively. Figure 2h depicts this comparison for Morocco.

The study by Rediat Takele [16], applied the ARIMA modelling approach to predict the prevalence patterns of coronavirus in four East African countries: Ethiopia, Djibouti, Sudan, and Somalia. The study utilized data from the reports of confirmed COVID-19 cases from March 13, 2020, to June 30, 2020, obtained from the official website of Johns Hopkins University. ARIMA models were used to predict the number of positive cases by the end of October 2020 under different scenarios. The scenarios were the worst-case scenario and the average-rate scenario. The Average-Rate Scenario assumes that average interventions and control measures are implemented to manage the spread of COVID-19. It projects the number of COVID-19 positive cases in the future based on this moderate level of intervention. The worst-case scenario represents a situation where there are insufficient interventions and control measures in place. It assumes a higher rate of transmission and projects a larger number of COVID-19 cases compared to the average-rate scenario. After calculating the RMSEs, it appeared that while the forecasts for Somalia, Djibouti and Sudan align well with the actual cases, the forecasts for Ethiopia show significant deviations. Figure 2i–l present the country-specific ARIMA forecasts.

According to a research article by Habtamu Legese Feyisa and Frezer Tilahun Tefera [17], an ARIMA model was used to analyze the spread of COVID-19 in Africa. The study collected daily confirmed new COVID-19 cases data from February 15, 2020, to October 16, 2020, from the official website of Our World in Data, to construct the ARIMA model and predict the trend of daily confirmed cases for the period between October 17, 2020, to November 16, 2020. The ARIMA models were constructed for Africa as a whole and for five African regions separately: East Africa, West Africa, Central Africa, North Africa, and Southern Africa. The study found that the ARIMA model forecasted values and the actual data had similar signs with slightly different sizes at the African level, and there were some deviations at the subregional level. The RMSE values for Africa as a continent and its five subregions (Northern Africa, Central or Middle Africa, Southern Africa, East Africa, and Western Africa) were approximately 3349.29, 1391.21, 373.00, 729.60, 247.28, and 186.62, respectively. Figure 2n–s illustrate forecasts for the African continent and five subregions.

The study by A. M. C. H. Attanayake and S. S. N. Perera [9] provides insights into the modelling and prediction of COVID-19 cases in eight countries, including South Africa. The data used in the study include accumulated COVID-19 cases from the first day of their presence until September 26, 2020. The paper reports that the Alpha-Sutte Indicator approach is appropriate in modelling cumulative COVID-19 cases in South Africa. The RMSE value calculated for the model is 1,361.96. This indicates that the Alpha-Sutte approach is effective in predicting COVID-19 cases in South Africa. However, the authors state that this model is suitable for short-term forecasting but may not perform well for long-term predictions. The reconstructed forecast for South Africa is shown in Figure 2m.

A study by Xu-Dong Liu et al. [18], proposed a prediction method that nests in-depth learning methods in the SIRV model to fit and predict the COVID-19 epidemic trend in Africa. The SIRV model refers to the Susceptible-Infected-Recovered-Dead-Vaccinated model, which is commonly used to model the spread of infectious diseases. The study aimed to address the unique transmission mode and low data quality, and incomplete data coverage of COVID-19 in Africa. The authors used data from worldometer to fit the COVID-19 transmission rate and trend from September 2021 to January 2022 for the top 15 African countries with the highest accumulative number of COVID-19 confirmed cases. The authors used non-autoregressive (NAR), LSTM, ARIMA models, Gaussian, and polynomial functions to predict the transmission rate $\beta$ in the next 7, 14, and 21 days. They then substituted the predicted transmission rate $\beta$s into the SIRV model to predict the number of COVID-19 active cases. In conclusion, the authors found that nesting the SIRV model with NAR, LSTM, ARIMA methods, etc., through functionalizing $\beta$, respectively, could obtain more accurate fitting and predicting results than these models/methods alone for the number of confirmed COVID-19 cases in Africa.

A summary of extracted details from each study is shown in Table 1.

One limitation of this study is the availability of a small number of data points for evaluating the performance of the proposed models. In certain cases, the limited data points reported posed a challenge in assessing the effectiveness and reliability of the models.

## 5. Discussion

This literature review examined the application of time-series analysis methods for forecasting the spread of COVID-19 in Africa. The findings reveal a variety of modelling approaches, each with distinct advantages and limitations. Among the reviewed studies, ARIMA and LSTM models emerged as the most widely applied. ARIMA models were often favored for their simplicity and efficiency in short-term forecasts, but were limited in capturing nonlinear dynamics and complex interactions. In contrast, deep learning models such as LSTM and GRU were effective at modelling complex temporal dependencies but demanded larger datasets and higher computational resources.

The comparative performance of these models varied across studies and contexts. For example, in South Africa, ARIMA outperformed LSTM in one study [2], whereas in Morocco, LSTM demonstrated superior accuracy compared to ARIMA [14]. These differences likely reflect variations in dataset length, stability, and reporting quality. Similarly, the poor alignment of ARIMA forecasts for Ethiopia compared to neighboring countries [15] illustrates the model's sensitivity to irregular reporting and missing values. Deep learning models such as LSTM and GRU achieved lower RMSEs when longer and more complete time series were available [10–12], but their performance degraded with smaller or noisier datasets. Compartmental approaches, such as the SIRV model, offered epidemiological interpretability but relied on parameter estimates (e.g., transmission rates) that were often difficult to obtain accurately in African contexts [17].

These findings highlight that forecasting accuracy depends not only on methodological design but also on the quality and completeness of the underlying data. In Africa, challenges such as underreporting, delays in data reporting, heterogeneous testing strategies, and limited healthcare infrastructure strongly influence model reliability. Models that assume consistent and complete data, such as ARIMA, may perform poorly in such contexts, while data-hungry approaches like LSTM risk overfitting. Hybrid models that combine statistical or mechanistic structures with machine learning components appear promising, as they may balance robustness with flexibility in handling imperfect data.

It is also important to emphasize that the RMSE values reported in this review are study-specific and should not be interpreted as directly comparable across studies. Different researchers drew from heterogeneous data sources such as Johns Hopkins University, Kaggle, Worldometer, and Our World in Data, which vary in reporting protocols, update frequency, and completeness. In cases where RMSE values were not reported, we recalculated them using Our World in Data to ensure internal consistency within this review. However, because some models were originally developed using alternative datasets, these recalculated RMSEs may not exactly reflect the authors' intended evaluation. Consequently, some of the observed performance differences may reflect dataset characteristics as much as methodological differences.

The choice of forecasting model should therefore be guided by the research objectives, data availability, computational resources, and the intended forecasting horizon. For short-term forecasting, ARIMA models remain a reliable choice due to their computational efficiency and low data requirements. For long-term forecasting, LSTM and other deep learning models have potential advantages, provided sufficient training data are available. Hybrid approaches that integrate time-series methods with epidemiological models, such as SIRV, represent a promising direction for improving predictive accuracy and robustness.

In conclusion, this review provides insights into the comparative strengths and weaknesses of different forecasting models applied in the African context. It also underscores the need for methodological standardization and improved data collection practices, as the heterogeneity of datasets currently limits the comparability of results across studies. Addressing these gaps will be essential for enhancing the effectiveness of forecasting models in supporting public health decision-making during future pandemics.

Table 2. Summary of the Results of Studies

| **Author, Year** | **Countries (Only in Africa)** | **Data Sources** | **Model(s)** | **Performance Metrics (Best Model)** | **Ref** |
|---|---|---|---|---|---|
| | | | | ***RMSE*** | |
| K.E. ArunKumar et al. 2021 | South Africa | John Hopkins University's COVID-19 database | GRU | 37428 | [11] |
| | | | LSTM | 2428 | |
| Nahla F. Omran et al. 2021 | Egypt | Novel Corona Virus 2019 Dataset taken from Kaggle | GRU | 670.30478 | [12] |
| | | | LSTM | 1067.02289 | |
| Mohamed Marzouk et al. 2021 | Egypt | https://flevy.com/coronavirus | LSTM | 1724 | [13] |
| Adewale F. Lukman et al. 2020 | South Africa [1], Nigeria [2], Ghana [3], and Egypt [4] | World Health Organization | ARIMA | 727.190 [1] | [14] |
| | | | | 50.082 [2] | |
| | | | | 122.331 [3] | |
| | | | | 293.516 [4] | |
| K.E. ArunKumar et al. 2022 | South Africa | John Hopkins University's COVID-19 database | ARIMA | 80225 | [3] |
| | | | SARIMA | 319775 | |
| | | | LSTM | 180225 | |
| | | | GRU | 230225 | |
| Mohamed Amine Rguibi et al. 2022 | Morocco | John Hopkins University's COVID-19 database, Moroccan Health Ministry | ARIMA | 1862.109 | [15] |
| | | | LSTM | 795.293 | |
| Rediat Takele, 2020 | Ethiopia[1], Djibouti[2], Sudan[3], and Somalia[4] | John Hopkins University's COVID-19 database | ARIMA (Average, Worst) | 26598, 12336 [1] | [16] |
| | | | | 3415, 6455 [2] | |
| | | | | 4642, 12645 [3] | |
| | | | | 47, 3775 [4] | |
| A. M. C. H. Attanayake and S. S. N. Perera, 2020 | South Africa | European Center for Disease Prevention and Control (ECDC) | Alpha-Sutte Indicator | 1361.96 | [9] |
| Habtamu Legese Feyisa and Frezer Tilahun Tefera, 2022 | Africa continent [1], East Africa [2], West Africa [3], Central Africa [4], North Africa [5], and Southern Africa [6] | Our World in Data COVID-19 database | ARIMA | 3349.29 [1] | [17] |
| | | | | 247.28 [2] | |
| | | | | 186.62 [3] | |
| | | | | 373.00 [4] | |
| | | | | 1391.21 [5] | |
| | | | | 729.60 [6] | |
| Xu-Dong Liu et al. (2023) | South Africa, Morocco, Tunisia, Libya, Egypt, Ethiopia, Kenya, Zambia, Algeria, Botswana, Nigeria, Zimbabwe, Mozambique, Uganda, and Ghana | Worldometer, Chinese Center for Disease Control and Prevention | ARIMA (alone, nested) | 45644, 28205 | [18] |
| | | | LSTM (alone, nested) | 44335, 18315 | |

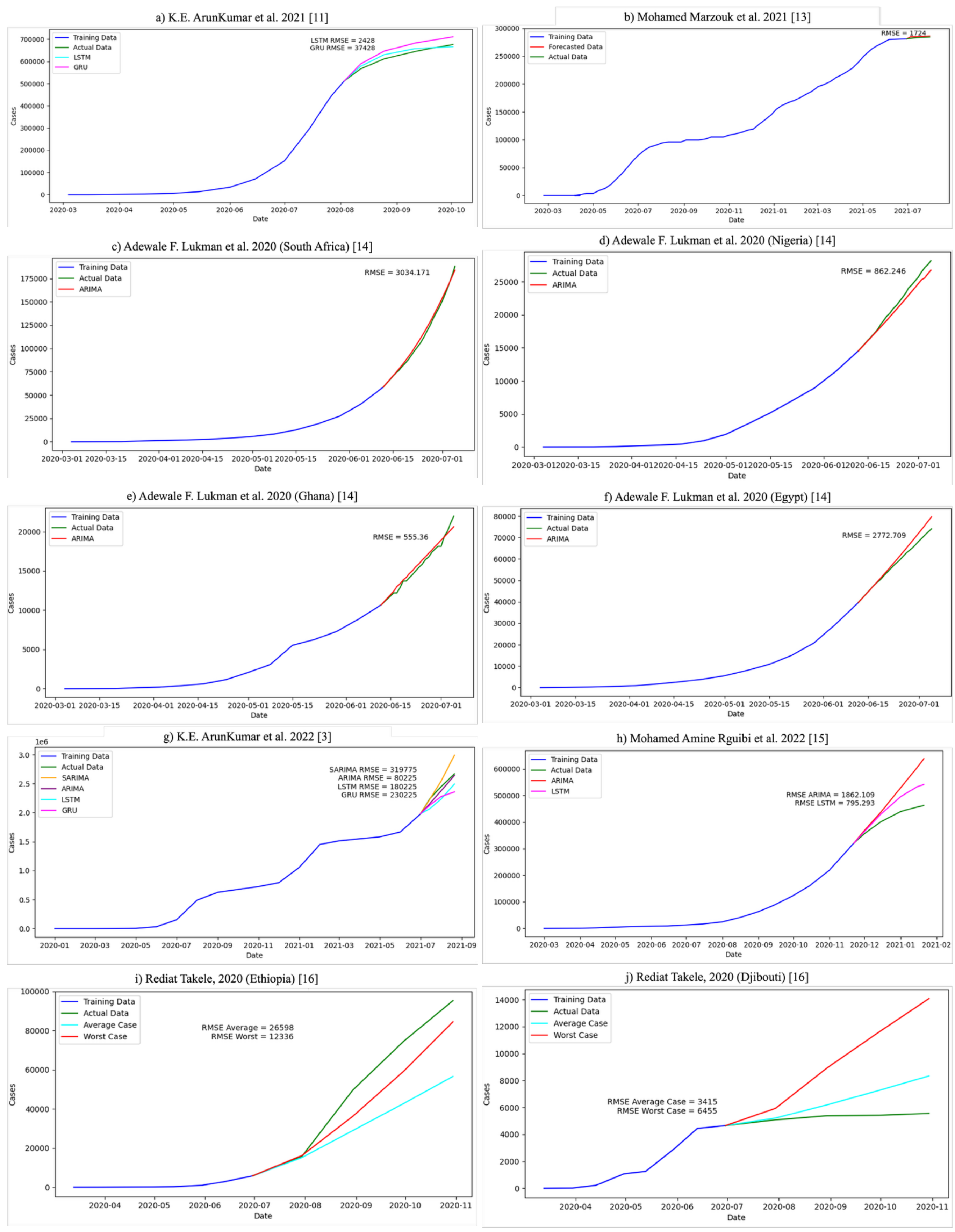

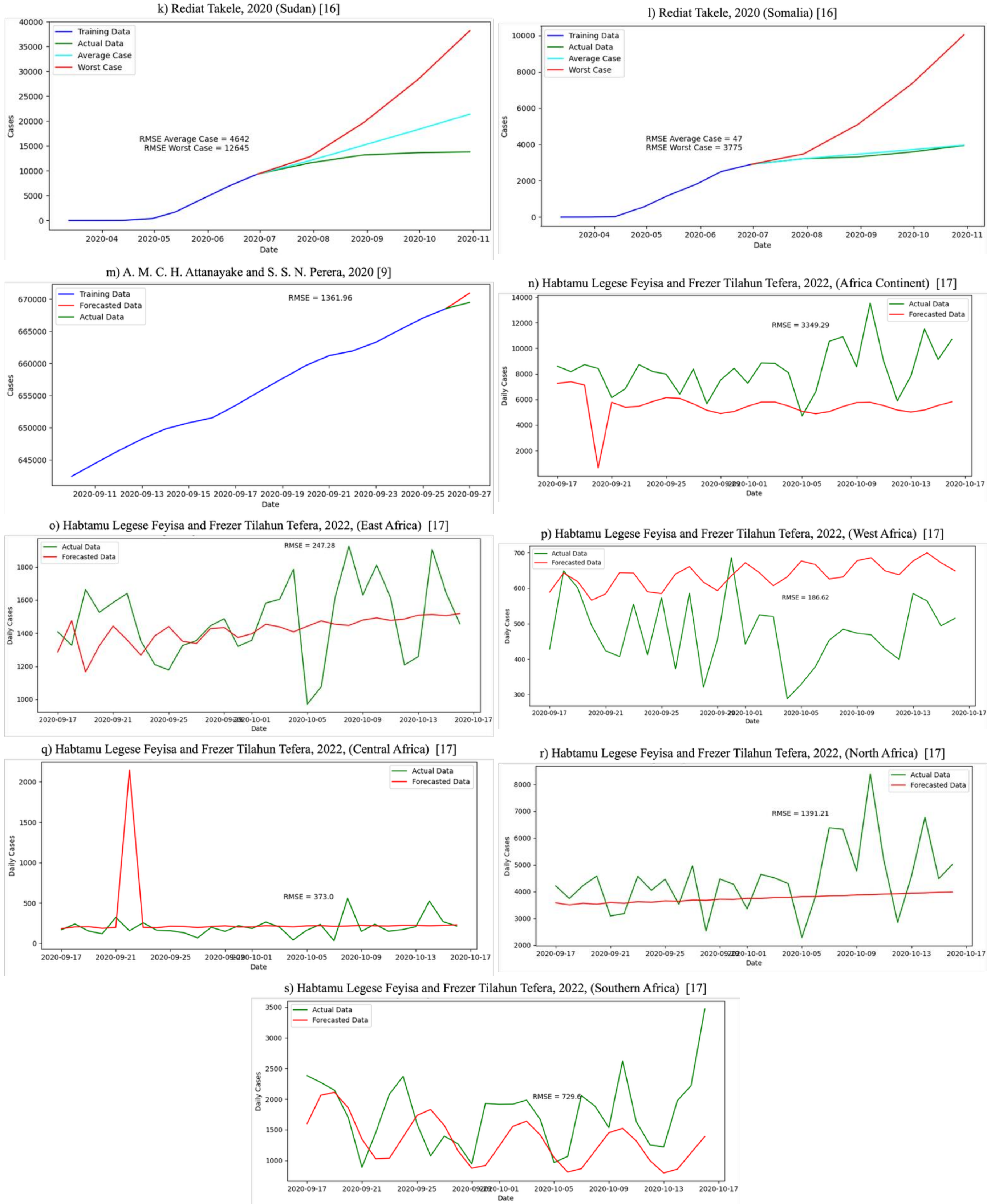

**Figure 2. Forecasting models of COVID-19 in Africa.** Charts (a–s) present model forecasts compared to actual case data, following the order of studies included in this review.

## References

[1] "World Health Organization COVID-19 dashboard," 2023. [Online]. Available: https://covid19.who.int/

[2] M. J. Page *et al.*, "The PRISMA 2020 statement: an updated guideline for reporting systematic reviews," *BMJ*, p. n71, Mar. 2021, doi: 10.1136/bmj.n71.

[3] K. E. ArunKumar, D. V. Kalaga, C. Mohan Sai Kumar, M. Kawaji, and T. M. Brenza, “Comparative analysis of Gated Recurrent Units (GRU), long Short-Term memory (LSTM) cells, autoregressive Integrated moving average (ARIMA), seasonal autoregressive Integrated moving average (SARIMA) for forecasting COVID-19 trends,” *Alexandria Engineering Journal*, vol. 61, no. 10, pp. 7585–7603, Oct. 2022, doi: 10.1016/j.aej.2022.01.011.

[4] S. Siami-Namini, N. Tavakoli, and A. S. Namin, “The Performance of LSTM and BiLSTM in Forecasting Time Series,” in *2019 IEEE International Conference on Big Data (Big Data)*, IEEE, Dec. 2019, pp. 3285–3292. doi: 10.1109/BigData47090.2019.9005997.

[5] K. Cho *et al.*, “Learning Phrase Representations using RNN Encoder-Decoder for Statistical Machine Translation.” doi: https://doi.org/10.48550/arXiv.1406.1078.

[6] J. Chung, C. Gulcehre, K. Cho, and Y. Bengio, “Empirical Evaluation of Gated Recurrent Neural Networks on Sequence Modeling,” Dec. 2014, doi: https://doi.org/10.48550/arXiv.1412.3555.

[7] A. S. Ahmar, A. Rahman, A. N. M. Arifin, and A. A. Ahmar, “Predicting movement of stock of ‘Y’ using Sutte Indicator,” *Cogent Economics & Finance*, vol. 5, no. 1, p. 1347123, Jan. 2017, doi: 10.1080/23322039.2017.1347123.

[8] A. Saleh Ahmar, “A Comparison of α-Sutte Indicator and ARIMA Methods in Renewable Energy Forecasting in Indonesia,” *International Journal of Engineering & Technology*, vol. 7, no. 1.6, p. 20, Jan. 2018, doi: 10.14419/ijet.v7i1.6.12319.

[9] A. M. C. H. Attanayake and S. S. N. Perera, “Forecasting COVID-19 Cases Using Alpha-Sutte Indicator: A Comparison with Autoregressive Integrated Moving Average (ARIMA) Method,” *Biomed Res Int*, vol. 2020, 2020, doi: 10.1155/2020/8850199.

[10] Edouard Mathieu *et al.*, “COVID-19 Pandemic,” *Our World in Data*, 2020, [Online]. Available: https://ourworldindata.org/coronavirus

[11] K. E. ArunKumar, D. V. Kalaga, C. M. S. Kumar, M. Kawaji, and T. M. Brenza, “Forecasting of COVID-19 using deep layer Recurrent Neural Networks (RNNs) with Gated Recurrent Units (GRUs) and Long Short-Term Memory (LSTM) cells,” *Chaos Solitons Fractals*, vol. 146, May 2021, doi: 10.1016/j.chaos.2021.110861.

[12] N. F. Omran, S. F. Abd-El Ghany, H. Saleh, A. A. Ali, A. Gumaei, and M. Al-Rakhami, “Applying Deep Learning Methods on Time-Series Data for Forecasting COVID-19 in Egypt, Kuwait, and Saudi Arabia,” *Complexity*, vol. 2021, 2021, doi: 10.1155/2021/6686745.

[13] M. Marzouk, N. Elshaboury, A. Abdel-Latif, and S. Azab, “Deep learning model for forecasting COVID-19 outbreak in Egypt,” *Process Safety and Environmental Protection*, vol. 153, pp. 363–375, Sep. 2021, doi: 10.1016/j.psep.2021.07.034.

[14] A. F. Lukman, R. I. Rauf, O. Abiodun, O. Oludoun, K. Ayinde, and R. O. Ogundokun, “COVID-19 prevalence estimation: Four most affected African countries,” *Infect Dis Model*, vol. 5, pp. 827–838, Jan. 2020, doi: 10.1016/j.idm.2020.10.002.

[15] M. A. Rguibi, N. Moussa, A. Madani, A. Aaroud, and K. Zine-dine, “Forecasting Covid-19 Transmission with ARIMA and LSTM Techniques in Morocco,” *SN Comput Sci*, vol. 3, no. 2, Mar. 2022, doi: 10.1007/s42979-022-01019-x.

[16] R. Takele, “Stochastic modelling for predicting COVID-19 prevalence in East Africa Countries,” *Infect Dis Model*, vol. 5, pp. 598–607, Jan. 2020, doi: 10.1016/j.idm.2020.08.005.

[17] H. Legese Feyisa and F. Tilahun Tefera, “The Validity of Autoregressive Integrated Moving Average Approach to Forecast the Spread of COVID-19 Pandemic in Africa,” *Discrete Dyn Nat Soc*, vol. 2022, 2022, doi: 10.1155/2022/2211512.

[18] X. D. Liu *et al.*, “Nesting the SIRV model with NAR, LSTM and statistical methods to fit and predict COVID-19 epidemic trend in Africa,” *BMC Public Health*, vol. 23, no. 1, Dec. 2023, doi: 10.1186/s12889-023-14992-6.

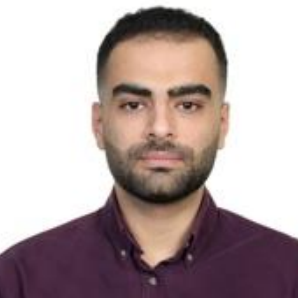

**Ali Ebadi** received his BSc degree in computer engineering from Persian Gulf University in 2023.

**Email:** ali.ebadi@mehr.pgu.ac.ir

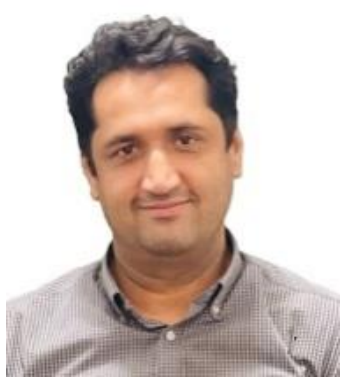

**Ebrahim Sahafizadeh** is an Assistant Professor in the Department of Computer Engineering at Persian Gulf University. His research interests include complex network analysis, computational epidemiology, soft security, and machine learning.

**Email:** sahafizadeh@pgu.ac.ir

**Paper Handling Data:**

Corresponding author: Ebrahim Sahafizadeh, Email: sahafizadeh@pgu.ac.ir
Affiliation of the corresponding author: Persian Gulf University